%% file: iclr_arxiv.tex
\documentclass{article}
\usepackage{iclr2027_conference_arxiv}
\iclrfinalcopy
\usepackage[T1]{fontenc}
\usepackage{times}
\usepackage{amsmath,amssymb,booktabs,graphicx,array,microtype}
\usepackage{float,placeins}
\usepackage{hyperref}
\usepackage{xurl}
\usepackage{algorithm}
\usepackage{algpseudocode}
\usepackage{titletoc}
\usepackage[dvipsnames]{xcolor}
\usepackage{annotate_equations}
\renewcommand{\eqnannotationtext}[1]{\sffamily\small #1\strut}

\hypersetup{hidelinks}
\newcommand{\method}{SteerAct}

\graphicspath{{figures/}}
\usepackage{colortbl}
\definecolor{linkc}{rgb}{0, 0.44, 0.74}
\definecolor{eqc}{rgb}{1, 0, 0}
\hypersetup{
  breaklinks=true,
  colorlinks=true,
  linkcolor=eqc,
  citecolor=linkc,
  urlcolor=eqc,
  pdftitle={SteerAct: Unlocking Action Models through Few-Shot Temporal Steering},
  pdfauthor={Anonymous Authors},
  pdfsubject={Research draft; experiments pending}
}

\title{SteerAct: Unlocking Action Models through Few-Shot Temporal Steering}

\author{%
\hspace*{-\tabcolsep}\makebox[\textwidth][c]{%
\begin{tabular}{c}
\textbf{Kaichen Zhou$^{1,2,\ast}$\quad
Zhiyan Li$^{2,\ast}$\quad
Chongkai Gao$^{3}$\quad
Yuhan Wang$^{4}$} \\[0.3em]
\textbf{Wenlong Huang$^{3}$\quad
Maggie Wang$^{3}$\quad
Fangneng Zhan$^{1}$\quad
Manling Li$^{5}$} \\[0.3em]
\textbf{Qianqian Wang$^{2}$\quad
Mac Schwager$^{3}$\quad
Zhuang Liu$^{6}$} \\[0.3em]
\textbf{Li Fei-Fei$^{3}$\quad
Mengyu Wang$^{2,\dagger}$\quad
Paul Pu Liang$^{1,\dagger}$} \\[0.6em]
\normalfont
$^{1}$Massachusetts Institute of Technology \quad
$^{2}$Harvard University \\[0.3em]
\normalfont
$^{3}$Stanford University \quad
$^{4}$Imperial College London \\[0.3em]
\normalfont
$^{5}$Northwestern University \quad
$^{6}$Princeton University \\[0.4em]
\normalfont
$^{\ast}$Equal contribution \quad
$^{\dagger}$Equal supervision
\end{tabular}}%
}

\begin{document}
\raggedbottom
\maketitle
\input{figures/figure1}
\input{section/0_abstract}
\input{section/1_introduction}
\input{section/2_related_work}

\input{section/3_methodology}
\input{section/4_experiments}

\input{section/5_conclusion}
\clearpage
{\small
\bibliography{iclr2027_conference}
\bibliographystyle{iclr2027_conference}
}
\clearpage
\appendix

\hypersetup{
  linkcolor=black,
  filecolor=red,
  urlcolor=red,
  pdftitle={Appendix},
  pdfauthor={Anonymous Authors}
}
\startcontents[appendix]
\section*{\textcolor{black}{Appendix}}
\makeatletter
\def\@currentlabel{Appendix}
\makeatother
\label{sec:appendix}
\setcounter{tocdepth}{2}
\printcontents[appendix]{}{1}{}

\input{section/6_appendix}
\end{document}

%% file: figures/figure1.tex
\begin{figure}[H]
\centering
\includegraphics[width=\linewidth]{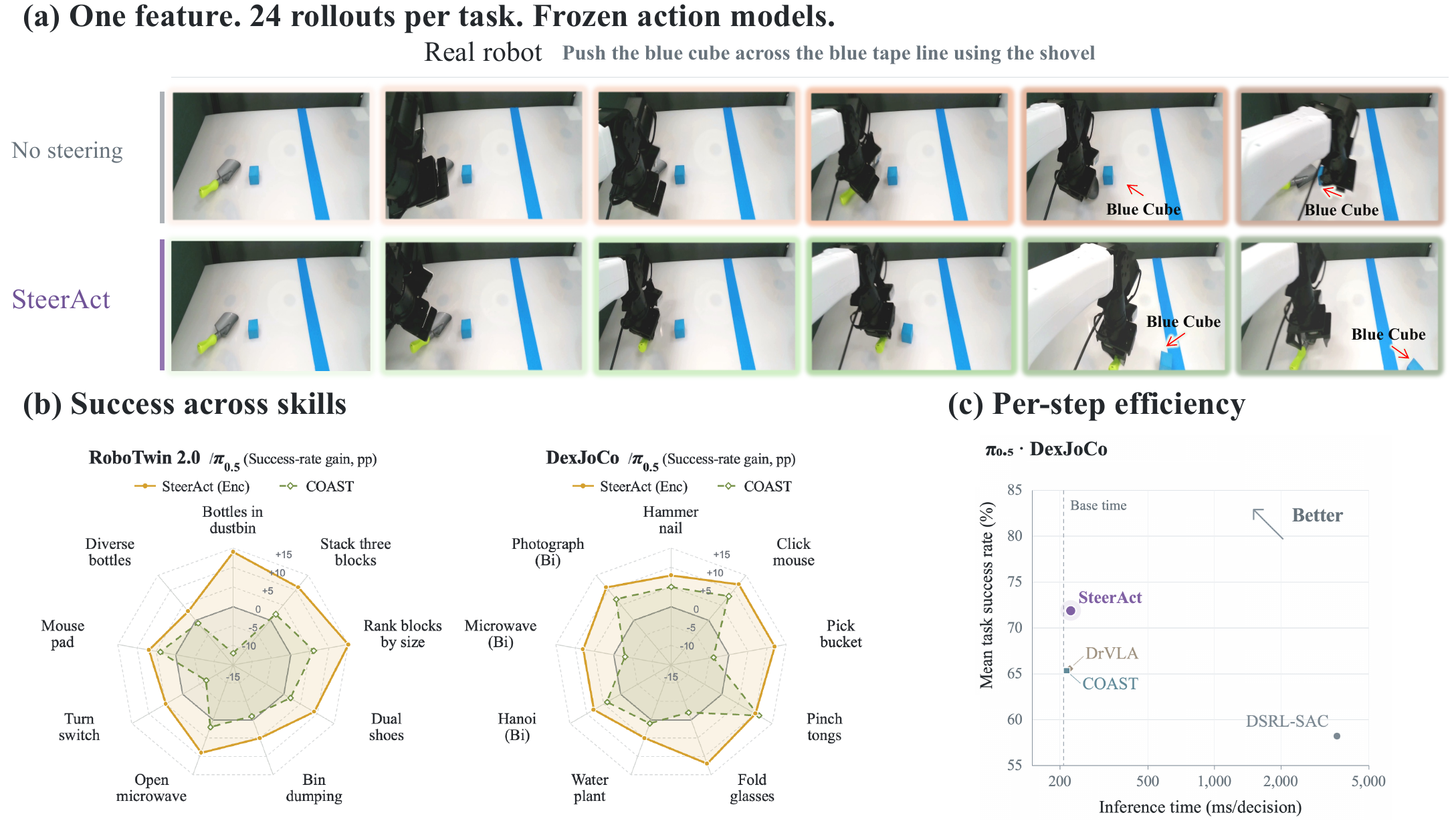}
\vspace{-13pt}
\caption{\textbf{Steering one feature at the right time.}
Using just 24 rollouts per task, SteerAct improves robot behavior by steering a single feature activation, without retraining the action model.
As the robot acts, SteerAct compares the patterns of how this feature activation changes with patterns from those existing rollouts to decide when to steer and how to steer.
\textbf{(a)} Unsteered and steered real-robot executions illustrate how \method{} improves task completion.
\textbf{(b)} SteerAct improves task success across all 18 evaluated RoboTwin~2.0 and DexJoCo tasks with $\pi_{0.5}$.
\textbf{(c)} On DexJoCo with $\pi_{0.5}$, SteerAct achieves the highest mean success among the evaluated methods.}
\vspace{-13pt}
\label{fig:hero}
\end{figure}

%% file: section/0_abstract.tex
\begin{abstract}
Improving pretrained robot action models at test time from only a few rollouts remains challenging.
Activation steering offers a lightweight approach by modifying feature activations while keeping model weights fixed.
However, effective steering requires identifying which features to modify and determining the timing and strength of modification.
We introduce \method{}, a few-shot temporal steering framework that uses a shared temporal descriptor of sparse-feature dynamics for offline feature selection and online control.
A pretrained sparse autoencoder provides candidate features, and our temporal descriptor summarizes the distribution of local activation-order patterns for each feature, enabling comparisons across rollouts without requiring timestep alignment. 
Before steering begins (Offline), we use successful and failed reference rollouts to select a single task-specific feature whose temporal descriptors best distinguish the two outcomes and construct its reference banks. 
During each new execution (Online), we compare the feature’s evolving temporal descriptor with these references to determine when to intervene and how strongly to increase or decrease its activation via a steering coefficient.
Using only 24 outcome-labeled rollouts per task, \method{} improves task success by approximately 10\% relative to unsteered models in evaluations spanning $\pi_{0.5}$, Fast-WAM, RoboTwin~2.0, DexJoCo, and real-world manipulation, without updating action-model weights.
Code and Demo can be found at: \href{https://steeract.github.io/}{Link}.
\newline
\textbf{Keywords:} Robot foundation models; Vision–language–action models (VLA); World-action models (WAM); Activation steering; Sparse autoencoders (SAEs); Few-shot adaptation.
\end{abstract}

%% file: section/1_introduction.tex
\section{Introduction}
\label{sec:introduction}

Pretrained robot action models, including vision--language--action (VLA) and world--action models (WAMs), provide a reusable foundation for manipulation~\citep{black2025pi05,yuan2026fastwam}.
Their capabilities, however, do not always translate into reliable task execution~\citep{zhang2026world,fei2025libero}.
When additional robot rollouts are limited, a central challenge is to turn the available execution experience into effective behavioral corrections~\citep{ball2023efficient,wagenmaker2025steering,gu2026safe,agia2024unpacking,xu2025can,park2026hide,mahato2026early}.
We therefore ask: \emph{Can we use a few successful and failed rollouts to identify and steer a single internal feature, enabling more reliable execution by a frozen action model?}

Inference-time steering offers a way to modify action generation while keeping the pretrained model frozen.
Existing approaches modify the initial noise of diffusion or flow policies~\citep{wagenmaker2025steering,tang2026improving}, or intervene on internal activations and
subspaces~\citep{haon2025mechanistic,swann2026sparse,miao2026coast}.
For feature-based steering, two connected questions arise: which features should be selected from a few rollouts, and when and how should their activations be modified during a new execution?
Addressing both requires a representation of feature activity that supports outcome-based feature selection and intervention decisions that adapt to the current rollout.

Our insight is that the temporal structure of feature activity can reveal outcome-related differences that activation magnitude alone misses.
A feature can have similar average activation in successful and failed rollouts yet differ in how its activity rises, falls, or fluctuates (Figure~\ref{fig:timing}).
We refer to the relative ordering of activation values within short windows as \emph{temporal activation patterns}.
Their distributions characterize local feature dynamics without requiring timestep alignment across executions, providing a common basis for identifying useful steering features and interpreting their activity during deployment.

Building on this insight, we introduce \method{}, a few-shot temporal steering framework that connects offline feature selection with online intervention through a shared temporal descriptor (Figure~\ref{fig:timing}).
A shared pretrained sparse autoencoder (SAE)~\citep{gao2025scaling,rajamanoharan2024jumping,lieberum2024gemma,marks2025sparse} provides candidate features and their decoder directions.
Before steering begins (offline), we construct success and failure reference banks from each candidate feature's temporal descriptors and select the feature with the strongest outcome separation.
During execution (online), we compare the observed feature history with the selected banks to determine when correction is needed and how to adjust the feature's activation, without updating the action model's weights.
Our contributions are summarized as follows:
\begin{itemize}

\item \textbf{Few-shot temporal steering of frozen action models.}
We introduce \method{}, a framework for improving pretrained robot action models from a few successful and failed rollouts through temporally guided interventions.
The action model remains frozen, and no auxiliary reinforcement-learning controller is trained.

\item \textbf{A shared temporal descriptor for feature selection and intervention.}
We represent sparse-feature dynamics through distributions of temporal activation patterns.
The same descriptor supports outcome-guided feature selection offline and reference-based control of intervention timing, and steering coefficient online.

\item \textbf{Evaluation across action models and manipulation
settings.}
Experiments with $\pi_{0.5}$ and Fast-WAM on RoboTwin~2.0
and DexJoCo, together with real-world manipulation,
demonstrate an approximately 10\% relative improvement
in task success over unsteered models using
24 outcome-labeled reference rollouts per task.
\end{itemize}

%% file: section/2_related_work.tex
\input{figures/figure2}

\section{Related Work}
\label{sec:related}

\subsection{Inference-time Steering}

Inference-time steering improves the behavior of a pretrained model during deployment without updating its backbone weights~\citep{liu2026vls, tan2026rl}. Candidate-selection methods sample multiple action proposals and rank them using a learned value function~\citep{nakamoto2024vgps}, an action verifier~\citep{kwok2025robomonkey}, a vision-language model reasoning over predicted outcomes~\citep{wu2025foresight}, or consistency with previous action chunks and likelihood of future plans~\citep{liu2025bidirectional}. Sampler methods instead modify the generative process. Building on classifier and classifier-free guidance for diffusion models~\citep{dhariwal2021diffusion, ho2021classifierfree}, these approaches bias sampling through user-provided guidance~\citep{wang2025itps} or objectives evaluated through learned dynamics~\citep{du2025dynaguide}. 
Others operate through the sampler's latent noise. 
DSRL treats the frozen sampler as an action transformation and learns a state-conditioned policy over its initial noise \citep{wagenmaker2025steering}, while 
flow reversal steering maps candidate actions back to the noise that produces them~\citep{tang2026improving}.   
These methods control action generation through noise selection or prediction-based guidance. \method{} instead performs temporal steering by directly modifying a sparse internal feature and using its activation history to determine when and how to intervene.
This requires hidden-state access but avoids learning an auxiliary RL controller or predicting future visual observations.

\subsection{Representation-level Steering and Sparse Autoencoders}
Representation-level steering modifies a model's hidden activations along directions associated with desired behavior~\citep{jin2026event,hong2026steering}.
Prior work has identified such directions through optimized latent vectors, contrastive activation differences, probing, and representation engineering~\citep{subramani2022extracting, turner2023steering,li2023inference, zou2023representation}.
SAEs provide a structured interface for representation-level control by decomposing model activations into a dictionary of sparse features~\citep{bricken2023monosemanticity,cunningham2024sparse}, whose decoder directions can provide candidate steering directions~\citep{templeton2024scaling}.
In robotics, mechanistic studies identify internal VLA directions associated with low-level behaviors such as motion direction and speed~\citep{haon2025mechanistic}. SAE-based VLA steering using contrastive action pairs has been used to construct latent steering directions for VLA control~\citep{khan2025controlling}, while individual SAE features have also been shown to encode interpretable motion primitives and semantic properties and to causally steer robot behavior \citep{swann2026sparse}. 
COAST instead uses successful and failed rollouts to fit contrastive conceptors that steer hidden states toward success-associated subspaces \citep{miao2026coast}. 
Its global and denoising-step-specific variants establish that outcome-guided activation steering need not rely on a single additive vector.
\method{} builds on prior representation-level steering methods, but uses the feature's activation history to control steering over the course of a rollout. From successful and failed reference rollouts, it selects a single sparse feature whose temporal activation patterns best distinguish the two outcomes. During deployment, it compares the observed feature history with the references to decide when and how to steer that feature.

%% file: figures/figure2.tex
\begin{figure}[t]
\centering
\includegraphics[width=0.9\linewidth]{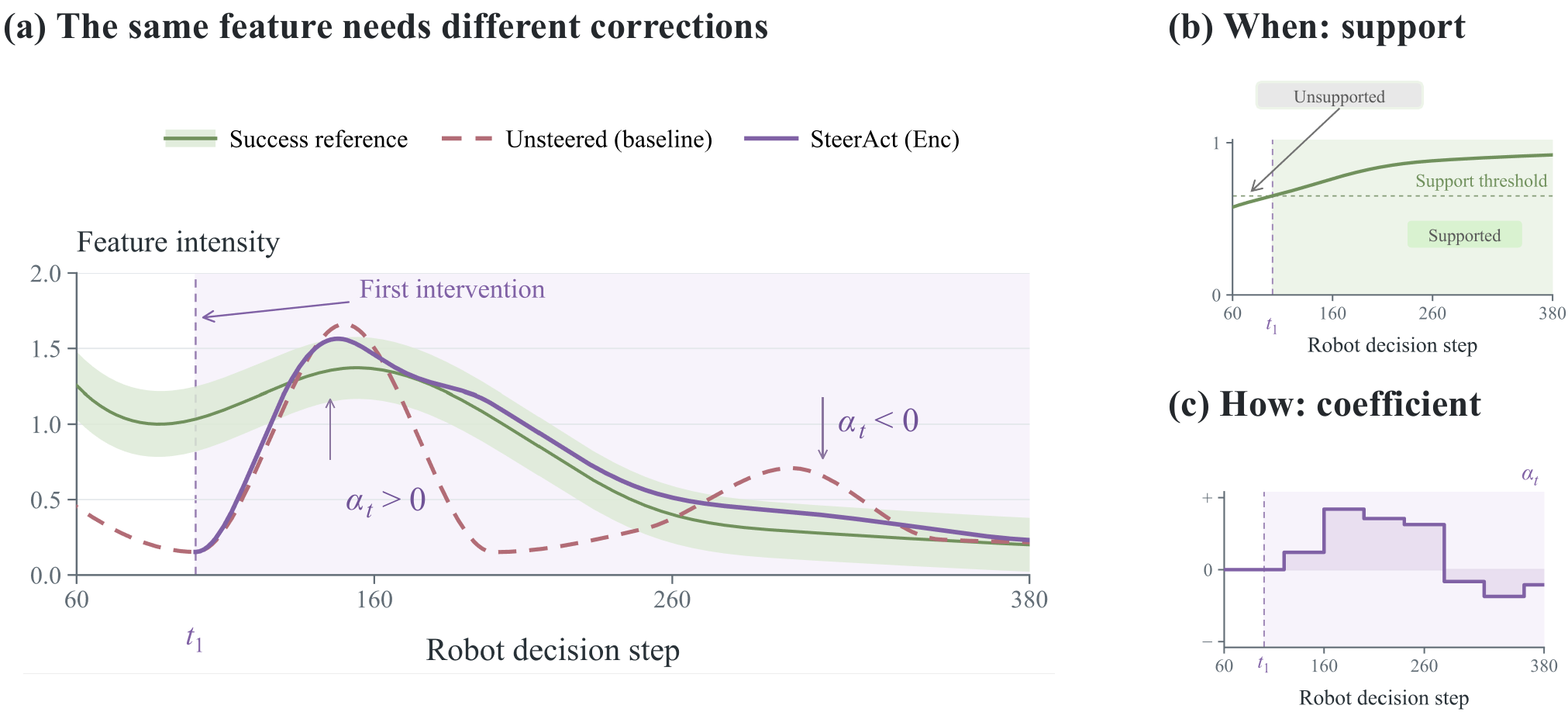}
\vspace{-0.3cm}
\caption{\textbf{Temporal evidence guides when and how to steer.}
\textbf{(a)} The selected feature's activation traces compare a successful reference, an unsteered rollout, and rollouts with adaptive steering.
\textbf{(b)} The temporal-support score measures whether the observed activation history of current rollout provides sufficient evidence for comparison with patterns from successful and failed reference rollouts.
\textbf{(c)} The steering coefficient $\alpha_t$ determines the direction and strength of the activation edit; $\alpha_t=0$ indicates no intervention at that decision.}
\vspace{-0.3cm}
\label{fig:timing}
\end{figure}

%% file: section/3_methodology.tex
\section{Methodology}
\label{sec:method}

\method{} is a few-shot temporal steering framework that uses a shared representation of feature dynamics for offline feature selection and online intervention (Figure~\ref{fig:method}).
A pretrained SAE decomposes the action model's hidden activations into sparse features, each associated with a decoder direction for intervention.
A temporal descriptor summarizes each feature's activation history, placing complete reference histories, observed execution prefixes, and candidate-edited histories in the same representation space.
Before steering begins (offline), differences between successful and failed references identify task-specific steering features.
During execution (online), comparisons in this space determine whether intervention is warranted and guide the steering coefficient.
The SAE is shared across tasks, and both its weights and the action model's weights remain frozen throughout.
The complete procedure is summarized in Algorithm~\ref{alg:steeract} in the appendix.

\subsection{Problem Setting}

We consider a frozen robot action model with access to its hidden activations at a fixed intervention site.
A pretrained sparse autoencoder (SAE), shared across tasks, encodes a hidden-state vector $\mathbf h\in\mathbb R^d$ into a sparse code $\mathbf z\in\mathbb R^K$ and reconstructs it through a linear decoder:
\begin{equation}
\mathbf z=E(\mathbf h),
\qquad
\widehat{\mathbf h}
=W_{\mathrm{dec}}\mathbf z+\mathbf b.
\label{eq:sae-representation}
\end{equation}
Here, $K$ is the number of SAE features.
In the reconstruction, each scalar activation $\mathbf z[f]$ weights a decoder column $\mathbf d_f=W_{\mathrm{dec}}[:,f]\in\mathbb R^d$.
Thus, a feature $f$ consists of an input-dependent activation and a fixed direction in the model's hidden space.
We use its activation history to guide intervention and its decoder direction to modify the original hidden state.

Given a few successful and failed reference rollouts, we select one task-specific feature $f^*$ offline and fix its steering direction $\mathbf d=\mathbf d_{f^*}$.
During a new rollout, let $\mathbf h_{t,j}$ denote the hidden-state vector of targeted token $j$ at robot decision $t$.
Here, $j$ indexes the tokens selected for intervention, while $t$ indexes robot decisions, not iterations within the action sampler.
Steering applies the same direction and steering coefficient to each targeted token:
\begin{equation}
\mathbf h'_{t,j}
=\mathbf h_{t,j}+\alpha_t\mathbf d.
\label{eq:intervention}
\end{equation}
Our online goal is to choose steering coefficient $\alpha_t$ using only the activation history observed so far.
The selected direction remains fixed, and neither the action model nor the SAE is updated.

\subsection{Temporal Descriptor}
\label{sec:temporal-descriptor}

Our representation captures the distribution of local temporal patterns in feature activity, rather than its average magnitude.
For feature $f$ in rollout $e$, we encode the measured hidden-state vectors with the SAE and define its intensity at decision $t$ as
\begin{equation}
y_{f,e,t}
=
\frac{1}{N_{e,t}}
\sum_{j=1}^{N_{e,t}}
\max\left\{
\left[E(\mathbf h_{e,t,j})\right]_f,0
\right\}.
\label{eq:feature-intensity}
\end{equation}
Here, $N_{e,t}$ is the number of measured tokens, and $[E(\mathbf h_{e,t,j})]_f$ is the activation of feature $f$ obtained by encoding token $j$'s hidden-state vector.
Thus, intensity measures feature activity through the SAE encoder, whereas steering modifies the hidden states from which this activity is measured.
For the selected feature in the current rollout, we write $y_t=y_{f^*,e,t}$.
We summarize the observed intensity history using ordinal patterns~\citep{bandt2002permutation}.
Omitting feature and rollout subscripts for readability, we form all available triples within $y_{1:t}$ at two spacings: $(y_i,y_{i+1},y_{i+2})$ and $(y_i,y_{i+2},y_{i+4})$.
Each triple is assigned to one of six patterns according to the relative ordering of its values.
Ties are resolved by temporal order, with earlier measurements ranked first.

For each spacing, we count the triples assigned to each pattern.
Dividing these six counts by the total number of triples gives the empirical pattern distribution.
We combine its six frequencies with three additional statistics: entropy, the fraction of triples containing ties, and an indicator of whether any triple is available.
These nine statistics per spacing form an 18-dimensional descriptor:
\begin{equation}
\mathbf v_{f,e,t}
=\phi(y_{f,e,1:t})\in\mathbb R^{18}.
\label{eq:ordinal}
\end{equation}
The descriptor is used to compare temporal behavior across rollouts.
We apply the same mapping $\phi$ to complete reference rollouts, observed execution prefixes, and histories modified by candidate corrections.
This connects offline feature selection with online steering without requiring timestep alignment across rollouts.

\input{figures/figure3}

\subsection{Offline Reference Construction and Feature Selection}
\label{sec:offline-steering}

Before steering begins (offline preparation), we use a few rollouts collected on the target task to identify a feature whose temporal patterns distinguish successful from failed executions.
For each candidate $f\in\mathcal C\subseteq\{1,\ldots,K\}$, we compute one descriptor from each complete reference rollout and group the descriptors by outcome (Success and Failure).
This produces a success bank $\mathcal B_f^S$ and a failure bank $\mathcal B_f^F$.
For outcomes $A,B\in\{S,F\}$, let $\mathbf v_{f,i}^A$
denote the $i$-th descriptor in bank $\mathcal B_f^A$.
We define
$
\bar d_f^{AB} =\operatorname{mean}_{i,j} \left\|\mathbf v_{f,i}^A-\mathbf v_{f,j}^B\right\|_2,
\label{eq:feature-distances}
$
where the mean includes all reference pairs, excluding $i=j$ when $A=B$.
Thus, $\bar d_f^{SF}$ measures between-outcome separation, while $\bar d_f^{SS}$ and $\bar d_f^{FF}$ measure within-outcome variability.
We select the feature with the greatest separation between these banks, accounting for variability within each outcome:
\begin{figure}[H]
\centering
\resizebox{\columnwidth}{!}{%
\begin{minipage}{\columnwidth}
\renewcommand{\eqnannotationtext}[1]{%
    \sffamily\small #1\strut}
\vspace*{1.7em}
\begin{equation}
\eqnmarkbox[Emerald]{sel-feature}{f^*} =
\arg\max_{
    \eqnmarkbox[BurntOrange]{candidate-feature}{f}
    \in \eqnmarkbox[OliveGreen]{candidate-set}{\mathcal C}}
\left(2\eqnmarkbox[NavyBlue]{cross-distance}{\bar d_f^{SF}}
- \eqnmarkbox[Plum]{success-distance}{\bar d_f^{SS}}
- \eqnmarkbox[WildStrawberry]{failure-distance}{\bar d_f^{FF}} \right).
\label{eq:selected-feature}
\end{equation}
\annotate[yshift=0.4em]{above,left}
{sel-feature}
{\shortstack[l]{Selected feature}}
\annotate[yshift=-0.4em]{below,left}
{candidate-feature}
{\shortstack[l]{Candidate feature}}
\annotate[yshift=-1.0em]{below,right}
{candidate-set}
{\shortstack[l]{Candidate feature set}}
\annotate[yshift=1.6em]{above,left}
{cross-distance}
{\shortstack[l]{Success--failure distance}}
\annotate[yshift=-0.4em]{below,right}
{success-distance}
{\shortstack[l]{Within-success distance}}
\annotate[yshift=0.4em]{above,right}
{failure-distance}
{\shortstack[l]{Within-failure distance}}
\vspace*{0.2em}
\end{minipage}%
}
\end{figure}
Selection therefore depends on differences in temporal patterns, not merely differences in average activation.
We retain the selected feature's decoder direction $\mathbf d$ and its reference banks,
$\mathcal B_S=\mathcal B_{f^*}^S$ and $\mathcal B_F=\mathcal B_{f^*}^F$.
All three remain fixed during deployment.
The same references used to select the feature now provide the basis for deciding when and how to steer it.

\subsection{Online Reference-Guided Steering}
\label{sec:online-steering}
During execution (Online), we monitor the selected feature and adapt only its steering coefficient $\alpha_t$.
At decision $t$, we append the current, unedited intensity $y_t$ to the recorded history and compute $\mathbf v_t=\phi(y_{1:t})$.
We first assess whether this history supports a meaningful reference comparison, then whether it resembles failure.
Only when both gates are open do we search for a correction.

\textbf{Support Gate.}
A short history can produce uncertain pattern frequencies even when its descriptor $\mathbf v_t$ is close to a reference.
In order to solve this problem, we propose a support score.
We draw $M$ plausible temporal descriptors from a Dirichlet model of the observed pattern counts to quantify uncertainty in the current activation history.
For each sampled descriptor, we count the reference descriptors in $\mathcal B_S\cup\mathcal B_F$ whose unscaled Euclidean distance from that sample is at most the offline-calibrated radius $\epsilon_{\mathrm{sup}}$.
A sample is supported when this count is at least $k_{\mathrm{sup}}$.
The support score $m_t$ is the fraction of the $M$ sampled descriptors that are supported.
The support check passes when $m_t\geq\tau_m$.
Sampling and calibration details are provided in Appendix~\ref{app:support-details}.

\textbf{Failure Gate.}
Passing the support gate doesn't mean that a correction is needed.
Let $D_S(\mathbf v)$ and $D_F(\mathbf v)$ denote the mean scaled Euclidean distances from descriptor $\mathbf v$ to its nearest $k_{\mathrm{sup}}$ references in the success and failure banks, respectively.
The failure check passes when $D_F(\mathbf v_t)<D_S(\mathbf v_t)$; equal distances do not pass.
To avoid rapid switching, each gate requires consecutive passing checks to open and consecutive failing checks to close.
Steering is enabled only when both gates are open; otherwise, $\alpha_t=0$.
The required check counts and gate-reset rules are specified in Appendix~\ref{app:distance-details}.

\textbf{Steering Coefficient.}
When both gates are open, we evaluate a finite coefficient grid $\mathcal A_t$ within fixed bounds, containing positive and negative values as well as zero.
Let $\mathbf w_{f^*}$ denote the selected feature's encoder weight vector, and let $\rho_t$ be the fraction of measured tokens on which that feature has positive activation.
Assuming a locally unchanged SAE active set, we select the coefficient whose predicted temporal descriptor is closest to the success bank:
\begin{figure}[H]
\centering
\resizebox{\columnwidth}{!}{%
\begin{minipage}{\columnwidth}
\renewcommand{\eqnannotationtext}[1]{%
    \sffamily\small #1\strut}
\vspace*{2.6em}
\begin{equation}
\begin{aligned}
\eqnmarkbox[Emerald]{alpha-selected}{\alpha_t} =
\arg\min_{\alpha\in
    \eqnmarkbox[OliveGreen]{alpha-grid}{\mathcal A_t}}
\eqnmarkbox[NavyBlue]{alpha-distance}{D_S}
\!\Bigl(
\eqnmarkbox[Plum]{alpha-descriptor}{\phi}
\!\Bigl(
\eqnmarkbox[BurntOrange]{alpha-history}
    {y_1,\ldots,y_{t-1}},
\eqnmarkbox[WildStrawberry]{alpha-prediction}{
    \max\!\left\{
        0,\,
        y_t+\alpha\rho_t
        \mathbf w_{f^*}^{\top}\mathbf d
    \right\}}
\Bigr)
\Bigr).
\end{aligned}
\label{eq:alpha-selection}
\end{equation}
\annotate[yshift=0.4em]{above,left}
{alpha-selected}
{\shortstack[l]{Steering coefficient}}

\annotate[yshift=-0.4em]{below,left}
{alpha-grid}
{\shortstack[l]{Candidate coefficients}}

\annotate[yshift=1.8em]{above,right}
{alpha-distance}
{\shortstack[l]{Success-bank distance}}

\annotate[yshift=-0.9em]{below,right}
{alpha-descriptor}
{\shortstack[l]{Temporal descriptor}}

\annotate[yshift=0.4em]{above,right}
{alpha-history}
{\shortstack[l]{Past intensities}}

\annotate[yshift=-0.5em]{below,right}
{alpha-prediction}
{\shortstack[l]{Predicted current intensity}}

\vspace{-0.2em}
\end{minipage}%
}
\end{figure}
\vspace{-5\baselineskip}
We apply the selected correction using Eq.~\eqref{eq:intervention}.
Approximation details and numerical safeguards are provided in Appendix~\ref{app:local-response-details}.

%% file: figures/figure3.tex
\begin{figure}[t]
\centering
\vspace{-0.8cm}
\includegraphics[width=\linewidth]{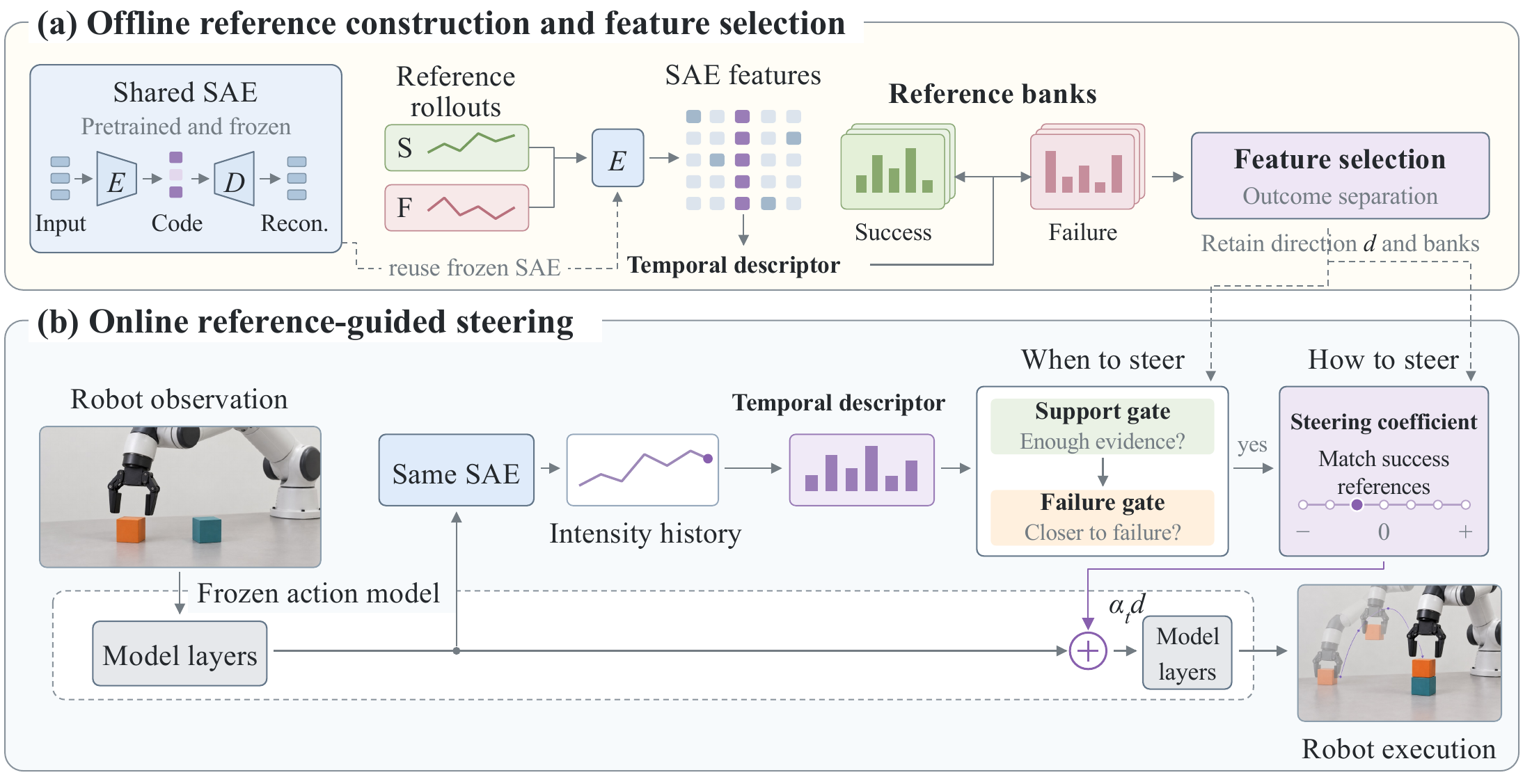}
\vspace{-0.8cm}
\caption{\textbf{\method{} overview.} Offline, an SAE fitted to frozen-policy
activations exposes candidate directions. Outcome-labeled rollouts determine a
feature and its success/failure banks of ordinal dynamics. Online, supported
failure-like prefixes enter a signed coefficient search; other prefixes retain
$\alpha_t=0$. The selected decoder direction modifies one configured residual
stream, after which the frozen policy generates actions.}
\vspace{-0.2cm}
\label{fig:method}
\end{figure}

%% file: section/4_experiments.tex
\section{Experiments}
\label{sec:experiments}

We evaluate whether outcome-conditioned temporal steering can improve closed-loop task success while keeping the underlying action policy frozen.
Our experiments address five questions:
(1) Does \method{} improve performance across different tasks and backbones (Section~\ref{sec:exp-main})?
(2) Does the method improve success in physical robot manipulation (Section~\ref{sec:exp-real})?
(3) Which temporal representations are most effective for outcome matching (Section~\ref{sec:exp-ablations})?
(4) Where should the intervention be applied (Section~\ref{sec:exp-ablations})?
(5) How does performance depend on the amount of source experience and the availability of successful references  (Section~\ref{sec:exp-ablations})?

\subsection{Experimental Setup}
\label{sec:exp-setup}

\textbf{Benchmarks and backbones.}
We evaluate on nine selected RoboTwin~2.0 tasks~\citep{chen2025robotwin} and nine selected DexJoCo tasks~\citep{wang2026dexjoco}, using both $\pi_{0.5}$ and Fast-WAM~\citep{yuan2026fastwam} on each benchmark (Section~\ref{sec:exp-main}).
This gives 36 simulated task--backbone evaluation settings. We additionally evaluate three real-robot tasks with $\pi_{0.5}$ (Section~\ref{sec:exp-real}).
\textbf{Setting.}
Unless otherwise stated, in simulation \method{} uses 24 outcome-labeled reference rollouts.
Experiments run on an NVIDIA RTX~6000 GPU.
\textbf{Baselines.}
\emph{Base} is the unmodified pretrained policy.
\emph{DSRL-SAC}~\citep{wagenmaker2025steering} trains a soft actor--critic controller over the initial noise supplied to a frozen generative action policy. For \emph{COAST}~\citep{miao2026coast}, we evaluate its positive-only variant, which constructs contrastive conceptor operators from successful and failed rollouts and uses them to steer hidden activations toward success-associated subspaces.
\emph{DrVLA}~\citep{swann2026sparse} uses sparse autoencoders to identify interpretable VLA features and intervenes along their directions.
We evaluate two variants of our method: \emph{\method{} (Enc)} intervenes in the backbone's encoder, whereas \emph{\method{} (Dec)} intervenes in its action expert, which applies steering at every denoising step whenever intervention is enabled.
\textbf{Metrics.}
Each method is evaluated with 200 held-out online rollouts per task. Success rates are binomial
proportions, so we test differences with two-proportion $z$-tests on counts pooled across tasks. All four aggregate gains over Base in Table~\ref{tab:robotwin} are significant ($n{=}1800$ per configuration; RoboTwin~2.0 with $\pi_{0.5}$: $z{=}5.9$, $p<10^{-8}$; the others at $p<0.05$). On the real robot ($n{=}90$ pooled), the 21.1~pp gain over Base is significant ($z{=}2.9$, $p{=}0.004$), whereas the 12.2~pp margin over DrVLA is not ($z{=}1.6$, $p{=}0.10$). Per-task differences at $n{=}200$ have a standard error of about 4~pp and are indicative only.

\input{tables/tab1}

\subsection{Evaluation Across Benchmarks, Embodiments, and Backbones}
\label{sec:exp-main}

We evaluate \method{} on gripper-based manipulation in RoboTwin~2.0 and dexterous-hand manipulation in DexJoCo, using $\pi_{0.5}$ and Fast-WAM as frozen backbones.
Each task has its own selected feature and reference banks, while the steering procedure remains unchanged across settings. Detailed experimental protocol and hyperparameters are listed in Appendix \ref{app:experimental-config}. Ablation studies supporting our method are shown in \ref{sec:exp-ablations}.

\textbf{Consistent aggregate gains.}
Table~\ref{tab:robotwin} shows that \method{} (Enc) achieves the highest reported mean success in all four benchmark--backbone settings.
Compared with Base, it gains 8.7 and 9.2 percentage points on RoboTwin~2.0, and 9.6 and 3.4 points on DexJoCo, with $\pi_{0.5}$ and Fast-WAM, respectively.
Gains are computed before rounding the task macro-averages.
These results support the applicability of the same procedure across the evaluated embodiments and architectures, although the magnitude of improvement varies.

\textbf{Broad task-level improvements.}
\method{} (Enc) improves over Base on all 18 RoboTwin~2.0 task--backbone pairs and on 34 of 36 pairs overall.
The exceptions are Click Mouse with Fast-WAM, which drops from 7\% to 5\%, and Hanoi Bi with Fast-WAM, which remains at 0\%.
Baseline effects are more mixed: DSRL-SAC improves aggregate success on RoboTwin~2.0 with $\pi_{0.5}$ but falls below Base on DexJoCo.
The advantage is therefore broad improvement over the unmodified policy, not dominance on every task;
individual baselines still achieve higher success on some tasks.

\textbf{Success gains with modest inference overhead.}
On RoboTwin~2.0, \method{} (Enc) adds 15.9~ms per decision with $\pi_{0.5}$ and 43.1~ms with Fast-WAM, corresponding to 7.6\% and 11.6\% overhead.
With $\pi_{0.5}$, its latency is close to DrVLA (223.8 versus 221.6~ms), while mean success is 6.2 points higher.
With Fast-WAM, it achieves higher mean success at slightly lower latency than DrVLA.
COAST is faster, but its mean success remains close to or below Base on these RoboTwin~2.0 evaluations.
Thus, \method{} offers a favorable success--latency trade-off without being the lowest-overhead method.

\textbf{Remaining backbone dependence.}
Aggregate improvements do not eliminate differences between the underlying models.
On DexJoCo, Fast-WAM improves from 23.3\% to 26.8\%, but remains well below the 71.9\% achieved with $\pi_{0.5}$.
Hanoi Bi also remains at 0\% across all reported Fast-WAM configurations.
Targeted activation steering can therefore improve execution reliability without closing the backbone-dependent performance gap.

\subsection{Real-Robot Evaluation}
\label{sec:exp-real}

\textbf{Experimental setup.}
We evaluate $\pi_{0.5}$ as a frozen backbone with I2RT YAM (2 Realsense D405 camera, 1 Realsense D435i camera) on three physical manipulation tasks: Folding Cloth, Pingpong Transport, and Push Cube.
Table~\ref{tab:real} compares Base, COAST, DrVLA, and the Enc and Dec variants of \method{}.
We report task-level success rates and their unweighted mean.

\input{tables/tab-3}

\input{figures/figure4}

\textbf{Consistent improvements on physical tasks.}
\method{} (Enc) achieves the highest reported success on all three tasks, reaching 52.2\% mean success compared with 31.1\% for Base and 40.0\% for DrVLA, the strongest evaluated baseline.
These correspond to improvements of 21.1 and 12.2 percentage points, respectively.
Gains over Base are 20.0 points on Folding Cloth, 23.3 on Pingpong Transport, and 20.0 on Push Cube.
The benefit therefore extends across cloth manipulation, object transport, and pushing, rather than being driven by a single physical task.

\textbf{Dependence on the intervention site.}
The two steering variants differ substantially on real hardware.
Enc outperforms Dec on every task, with mean success of 52.2\% versus 33.3\%.
Dec remains close to Base overall and provides no improvement on Push Cube.
These results favor the action model's encoder-branch over action-expert steering for the evaluated physical tasks, showing that a frozen policy's response to steering depends on where the intervention is applied.

\subsection{Ablation Studies}
\label{sec:exp-ablations}

We examine representation, intervention site, reference data, and steering choices on RoboTwin~2.0 with $\pi_{0.5}$.
VLABench provides complementary comparisons of timing and gating.

\textbf{Temporal descriptors improve outcome matching.}
Table~\ref{tab:ablations-1}(a) compares different descriptors.
The proposed descriptor achieves 77.7\% mean success, exceeding autocorrelation, the strongest alternative descriptor, by 3.8 percentage points.
Raw NN and DTW reach 72.2\% and 71.7\%, respectively: direct history matching and explicit temporal alignment do not recover the same aggregate performance.
These results favor our descriptor under the shared gating scheme (Appendix~\ref{app:ablation-variants}).

\input{tables/tab3}

\textbf{More reference rollouts do not guarantee higher success.}
With 12 total reference rollouts per task,
\method{} reaches 73.6\% mean success, 4.6 points above Base (Table~\ref{tab:ablations-1}(b)).
Increasing the budget to 48 improves all nine tasks and raises the mean to 77.3\%.
However, the default 24-rollout configuration already achieves 77.7\%, and increasing from 48 to 96 leaves the mean unchanged, with four task improvements, four declines, and one tie.
Thus, the reported results demonstrate effective few-shot adaptation, but not a monotonic benefit from enlarging the reference set.

\textbf{Additional successful references yield diminishing
aggregate gains.}
Table~\ref{tab:ablations-1}(c) varies the number of successful reference rollouts.
Increasing this count from 8 to 24 raises mean success from 74.4\% to 77.6\%, a gain of 3.1 points.
Increasing it further to 48 adds only 0.3 points, reaching 77.9\%, with mixed task-level changes.
These settings count successful episodes rather than all source episodes and are therefore not directly
comparable to the default budget of 24 total rollouts.
In particular, failed references must also be available for outcome-conditioned selection and steering.

\textbf{A single feature is effective, but additional features improve performance.}
Table~\ref{tab:ablations-1}(d) shows that steering the top two and four features achieves 79.6\% and 80.0\% mean success, respectively, compared with 77.7\% for the single-feature configuration.
The corresponding gains are 1.9 and 2.3 points. These results establish that one feature is sufficient
for substantial improvement over Base, not that one feature maximizes performance among the tested variants.
The same group tests coefficient adaptation while retaining the intervention gates.
Fixed coefficients of $\alpha=1$ and $\alpha=2$ achieve 75.8\% and 75.7\%, respectively, below the adaptive single-feature configuration by 1.9 and 2.0 points.
Fixed corrections remain competitive on individual tasks, but neither matches the adaptive variant's aggregate success.
This comparison supports choosing the correction during execution rather than prescribing either tested constant.

\textbf{Both gating components contribute to aggregate success.}
In the VLABench benchmark~\citep{zhang2025vlabench} (Table~\ref{tab:timing-summary}) , the fully gated variant achieves 36.13\% mean success and improves over the frozen policy on all eight tasks.
Removing the support gate reduces the mean by 13.00 percentage points, while removing the failure gate reduces it by 6.25 points.
Disabling both gates yields 22.38\%, below the frozen policy's 23.63\%.
Neither remaining gate alone reproduces the full variant's aggregate benefit, supporting selective rather than unconditional intervention in this setting.
Moreover, steering with our support gate achieves higher success than fixed-delay baselines that enable intervention only after 10 or 20 decision steps. This suggests that adapting intervention timing to the observed activation history is more effective than using these predetermined delays.

\input{tables/tab4}

%% file: tables/tab1.tex
\begin{table}[tbp]
\centering
\vspace{-0.8cm}
\caption{\textbf{SteerAct improves success across backbones
and manipulation benchmarks.}
Among the evaluated methods, \method{} (Enc) achieves the highest mean success in all four benchmark--backbone settings with modest inference overhead.
Success rates are reported in percent ($\uparrow$); each Mean averages nine tasks.
Time reports per-decision inference latency (ms, $\downarrow$), shared across datasets.
Enc/Dec denote encoder-branch/action-expert steering.
Gray highlights \method{} (Enc); \textbf{bold}/\underline{underline} mark the best/second-best distinct success rates within each backbone, including ties.}
\label{tab:robotwin}
\footnotesize
\setlength{\tabcolsep}{2.5pt}
\renewcommand{\arraystretch}{1.12}
\resizebox{\linewidth}{!}{%
\begin{tabular}{@{}l*{10}{c}
                   @{\hspace{4pt}}|@{\hspace{4pt}}
                   *{10}{c}
                   @{\hspace{8pt}}r@{}}
\toprule
& \multicolumn{10}{c}{\textbf{RoboTwin~2.0}}
& \multicolumn{10}{c}{\textbf{DexJoCo}}
& \\
\cmidrule(lr){2-11}
\cmidrule(lr){12-21}
Method
& \shortstack{Bottles\\in Dustbin}
& \shortstack{Stack Three\\Blocks}
& \shortstack{Rank Blocks\\by Size}
& \shortstack{Dual\\Shoes}
& \shortstack{Dump Bin\\into Big Bin}
& \shortstack{Open\\Microwave}
& \shortstack{Turn\\Switch}
& \shortstack{Mouse\\Pad}
& \shortstack{Diverse\\Bottles}
& Mean
& \shortstack{Hammer\\Nail}
& \shortstack{Click\\Mouse}
& \shortstack{Pick\\Bucket}
& \shortstack{Pinch\\Tongs}
& \shortstack{Fold\\Glasses}
& \shortstack{Water\\Plant}
& \shortstack{Hanoi\\Bi}
& \shortstack{Microwave\\Bi}
& \shortstack{Photo\\Bi}
& Mean
& \shortstack{Time $\downarrow$\\(ms/decision)} \\
\midrule

\multicolumn{22}{@{}l}{
\textit{Backbone: $\pi_{0.5}$}} \\
\midrule

Base
& 74 & 72 & 43 & 73 & 89 & 84 & 47 & 62 & 77
& 69.0
& 82 & 67 & 83 & 22 & 71 & 92 & 15 & 77 & 52
& 62.3
& 207.9 \\

DSRL-SAC
& \underline{80} & 78 & 40 & 57 & 49
& \textbf{100} & \textbf{65} & \textbf{71} & \textbf{100}
& 71.1
& 80 & 60 & 70 & 18 & 74 & \underline{96} & 10 & 69 & 47
& 58.2
& 3582.7 \\

DrVLA
& 62 & \textbf{84} & \textbf{60} & \textbf{86}
& 72 & 91 & 48 & 68 & 72
& \underline{71.4}
& 85 & \underline{76} & \underline{87} & 26
& 68 & 94 & 15 & 84 & 55
& 65.6
& 221.6 \\

COAST
& 62 & 74 & 49 & 75 & 88 & 86 & 40 & 66 & 76
& 68.4
& \underline{87} & 75 & 79 & \textbf{33}
& 69 & 93 & \underline{19} & 74 & \underline{59}
& 65.3
& 214.4 \\

\method{} (Dec)
& 79 & \underline{83} & 39 & 81 & \underline{92}
& 87 & 44 & 67 & 66
& 70.9
& 79 & 74 & 75 & 21 & \textbf{87}
& \underline{96} & 17 & \textbf{91} & \underline{59}
& \underline{66.6}
& 278.1 \\

\rowcolor[gray]{0.93}
\method{} (Enc)
& \textbf{88} & \underline{83} & \underline{58}
& \underline{82} & \textbf{94} & \underline{93}
& \underline{52} & \underline{69} & \underline{80}
& \textbf{77.7}
& \textbf{90} & \textbf{79} & \textbf{95} & \underline{32}
& \underline{83} & \textbf{97} & \textbf{23}
& \underline{85} & \textbf{63}
& \textbf{71.9}
& 223.8 \\

\midrule
\multicolumn{22}{@{}l}{
\textit{Backbone: Fast-WAM}} \\
\midrule

Base
& 72 & 77 & 63 & 71 & \underline{91}
& 31 & 53 & 87 & 79
& 69.3
& 22 & \textbf{7} & 70 & 12 & \underline{4}
& 52 & \textbf{0} & 19 & 24
& 23.3
& 372.6 \\

DrVLA
& \underline{78} & 72 & \textbf{76} & 73
& \textbf{94} & \underline{52} & \underline{63}
& 74 & \textbf{84}
& 74.0
& \underline{25} & 3 & \underline{73} & 17 & 2
& \textbf{63} & \textbf{0} & \textbf{25} & 22
& \underline{25.6}
& 419.2 \\

COAST
& 64 & 72 & 74 & 77 & 84
& \underline{52} & 59 & 78 & 66
& 69.6
& 21 & \textbf{7} & 66 & 15 & \textbf{5}
& 55 & \textbf{0} & 21 & \underline{25}
& 23.9
& 387.5 \\

\method{} (Dec)
& \textbf{84} & \textbf{79} & \underline{75}
& \underline{81} & \underline{91} & \textbf{54}
& 60 & \underline{90} & 80
& \underline{77.1}
& 24 & \underline{5} & 63 & \textbf{21} & \textbf{5}
& 57 & \textbf{0} & \textbf{25} & 17
& 24.1
& 423.9 \\

\rowcolor[gray]{0.93}
\method{} (Enc)
& \textbf{84} & \underline{78} & 72
& \textbf{85} & \textbf{94} & 43 & \textbf{73}
& \textbf{95} & \underline{83}
& \textbf{78.6}
& \textbf{27} & \underline{5} & \textbf{76} & \underline{19}
& \textbf{5} & \underline{59} & \textbf{0}
& \underline{23} & \textbf{27}
& \textbf{26.8}
& 415.7 \\

\bottomrule
\end{tabular}%
}
\vspace{-16pt}
\end{table}

%% file: tables/tab-3.tex
\begin{table}[tbp]
\centering
\vspace{-0.6cm}
\caption{\textbf{SteerAct improves success on all three
real-robot tasks.}
With $\pi_{0.5}$ frozen, \method{} (Enc) achieves the
highest success on every evaluated task, reaching
52.2\% average success compared with 31.1\% for Base
and 40.0\% for the strongest baseline, DrVLA.
Values are success rates (\%, $\uparrow$).
Enc/Dec denote encoder-branch/action-expert steering.
Gray highlights \method{} (Enc);
\textbf{bold}/\underline{underline} mark the best/second-best
distinct results in each row.}
\label{tab:real}

\small
\setlength{\tabcolsep}{6pt}
\renewcommand{\arraystretch}{1.12}

\begin{tabular}{
    @{}l
    *{4}{c}
    >{\columncolor[gray]{0.93}[\tabcolsep][0pt]}c
    @{}
}
\toprule
Task
& Base
& COAST
& DrVLA
& \shortstack{\method{}\\Dec}
& \shortstack{\method{}\\Enc} \\
\midrule

Folding Cloth
& 33.3\%
& 26.7\%
& \underline{40.0\%}
& 36.7\%
& \textbf{53.3\%} \\

Cup Ping-Pong
& 40.0\%
& 36.7\%
& \underline{56.7\%}
& 43.3\%
& \textbf{63.3\%} \\

Push Cube
& 20.0\%
& 0\%
& \underline{23.3\%}
& 20.0\%
& \textbf{40.0\%} \\

\midrule
\textbf{Average}
& 31.1\%
& 21.1\%
& \underline{40.0\%}
& 33.3\%
& \textbf{52.2\%} \\

\bottomrule
\end{tabular}
\end{table}

%% file: figures/figure4.tex
\begin{figure}[t]
\centering
\vspace{-8pt}
\includegraphics[width=\linewidth]{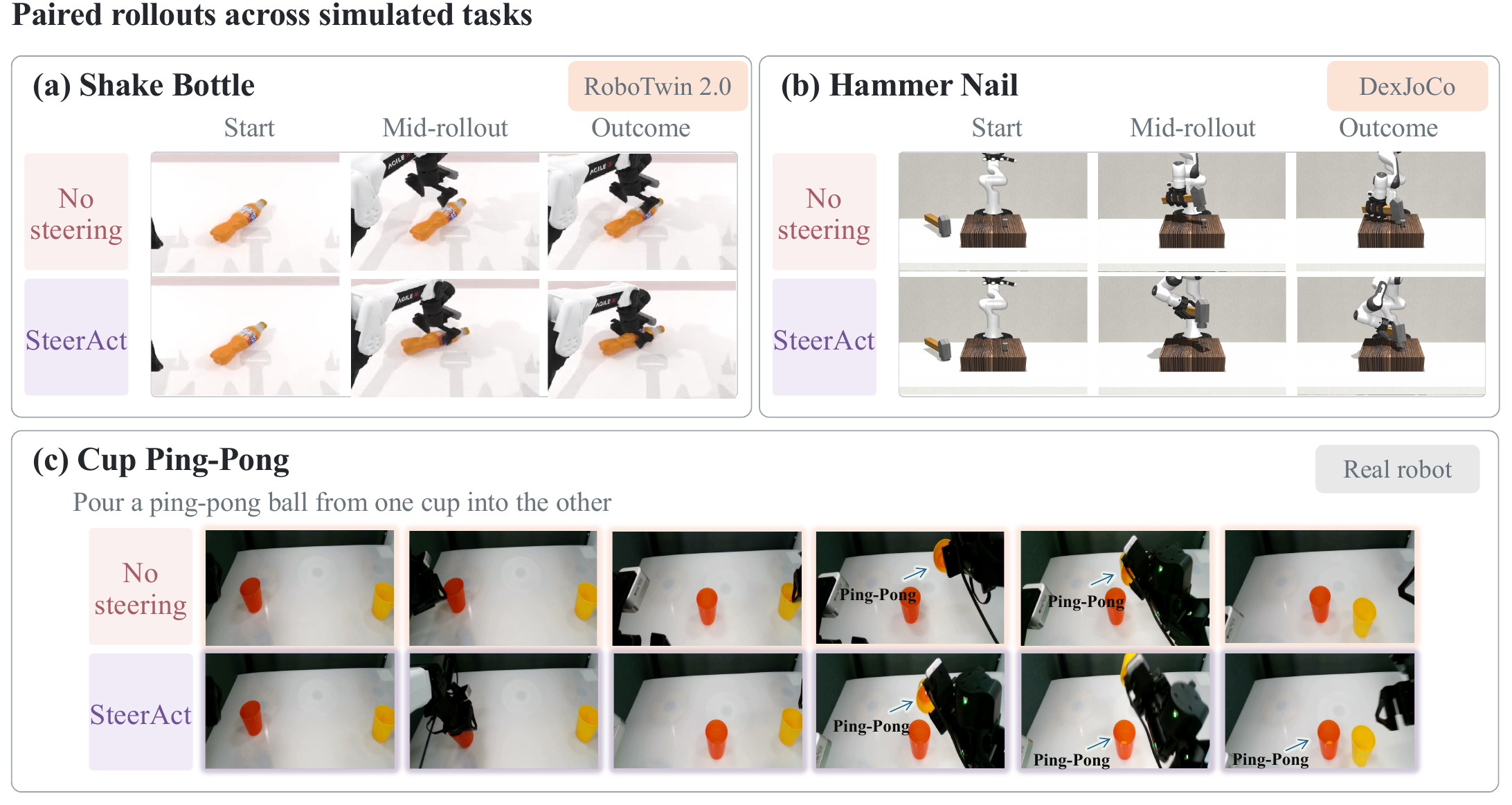}
\vspace{-18pt}
\caption{\textbf{Single-feature steering improves task completion
across manipulation settings.}
Representative unsteered and steered rollouts on RoboTwin~2.0, DexJoCo, and real robots illustrate how \method{} corrects execution errors and enables successful task completion while keeping the underlying action model frozen.}
\vspace{-6pt}
\label{fig:real}
\end{figure}

%% file: tables/tab3.tex
\begin{table}[tbp]
\centering
\vspace{-0.2cm}
\caption{Ablations on RoboTwin~2.0 with $\pi_{0.5}$.
Values are mean success rates (\%) across nine tasks.
Gray marks the default single-feature Enc variant
(24 total source rollouts).
Panels (b)--(d) use Enc; (c) counts successful source rollouts only.
Full results are in Appendix Table~\ref{tab:ablations}.}
\label{tab:ablations-1}
\begingroup
\small
\setlength{\tabcolsep}{2.5pt}
\renewcommand{\arraystretch}{1.15}

{\raggedright
\textbf{(a) Representation and intervention site}\par}
\vspace{3pt}
\begin{tabular*}{\linewidth}{@{\extracolsep{\fill}}
l*{8}{c}
>{\columncolor[gray]{0.94}[2.5pt][0pt]}c@{}}
\toprule
Method
& Base & Marginals & Increments & Events & Autocorr.
& Raw NN & Raw DTW & Dec & \textbf{Enc} \\
\midrule
Success
& 69.0 & 71.1 & 70.0 & 70.1 & \underline{73.9}
& 72.2 & 71.7 & 70.9 & \textbf{77.7} \\
\bottomrule
\end{tabular*}

\par\vspace{8pt}\noindent
\begin{minipage}[t]{0.29\linewidth}
\vspace{0pt}
\textbf{(b) Total rollouts}\par
\vspace{3pt}
\begin{tabular*}{\linewidth}{@{\extracolsep{\fill}}lr@{}}
\toprule
Rollouts & Success \\
\midrule
12 & \underline{73.6} \\
48 & \textbf{77.3} \\
96 & \textbf{77.3} \\
\bottomrule
\end{tabular*}
\end{minipage}\hfill
\begin{minipage}[t]{0.29\linewidth}
\vspace{0pt}
\textbf{(c) Successful rollouts}\par
\vspace{3pt}
\begin{tabular*}{\linewidth}{@{\extracolsep{\fill}}lr@{}}
\toprule
Rollouts & Success \\
\midrule
8  & 74.4 \\
24 & \underline{77.6} \\
48 & \textbf{77.9} \\
\bottomrule
\end{tabular*}
\end{minipage}\hfill
\begin{minipage}[t]{0.34\linewidth}
\vspace{0pt}
\textbf{(d) Steering choices}\par
\vspace{3pt}
\begin{tabular*}{\linewidth}{@{\extracolsep{\fill}}lr@{}}
\toprule
Setting & Success \\
\midrule
Top-$2$ features  & \underline{79.6} \\
Top-$4$ features  & \textbf{80.0} \\
Fixed $\alpha=1$ & 75.8 \\
Fixed $\alpha=2$ & 75.7 \\
\bottomrule
\end{tabular*}
\end{minipage}
\par
\endgroup
\end{table}

%% file: tables/tab4.tex
\begin{table}[tbp]
\centering
\vspace{-0.6cm}
\caption{Mean success rates (\%, $\uparrow$) across eight
VLABench tasks, with 200 evaluation episodes per task and setting ($\pi_{0.5}$ with \method{}(Enc)).
Full results are in Appendix Table~\ref{tab:pilot-timing}.}
\label{tab:timing-summary}
\footnotesize
\setlength{\tabcolsep}{6pt}
\renewcommand{\arraystretch}{1.15}
\resizebox{0.8\linewidth}{!}{%
\begin{tabular}{@{}l*{6}{c}
>{\columncolor[gray]{0.93}[\tabcolsep][0pt]}c@{}}
\toprule
Setting
& Frozen
& Delay 10
& Delay 20
& \shortstack{Without\\support gate}
& \shortstack{Without\\failure gate}
& \shortstack{Without\\gates}
& Ours \\
\midrule
Mean
& 23.63
& 24.63
& 24.88
& 23.13
& \underline{29.88}
& 22.38
& \textbf{36.13} \\
\bottomrule
\end{tabular}%
}
\vspace{-12pt}
\end{table}

%% file: section/5_conclusion.tex
\section{Conclusion}
We introduced \method{}, a few-shot framework for steering
frozen action models through a single sparse feature per task.
Temporal reference matching connects feature selection,
intervention timing, and signed correction, while
uncertainty-aware gating withholds unsupported interventions.
Experiments on RoboTwin~2.0 and DexJoCo show mean success
gains across $\pi_{0.5}$ and Fast-WAM, spanning gripper
and dexterous-hand manipulation.
Remaining regressions and backbone-dependent performance
highlight the limits of steering alone.
Together, these findings show that a single internal feature
can provide a useful adaptation interface when its intervention
is guided by the unfolding execution, without updating
model weights.

%% file: section/6_appendix.tex
\clearpage
\section{Foundation Action Models}
Vision-language-action (VLA) and world-action models (WAMs) leverage large-scale pretraining and diverse robot experience to support general-purpose manipulation~\citep{brohan2023rt2, kim2024openvla, black2025pi05, ye2026dreamzero, yuan2026fastwam,black2024pi_0,pertsch2025fast,kim2025fine,team2024octo,liu2025rdt,bjorck2025gr00t,team2025gemini,shukor2025smolvla,o2024open,zhou2026gem}.
While these models provide generalizable pretrained policies, their performance can remain unreliable on target tasks, motivating the need to improve their performance from limited additional experience.
Our work addresses adaptation after such representations have been learned. 
Rather than introducing a new action-model architecture or updating its parameters, \method{} uses outcome-labeled rollouts to guide interventions in the hidden activations of a frozen model to improve its performance on target tasks.

\input{tables/tab-7}
\input{tables/tab-8}

\input{figures/algorithm}

\section{Additional Details of Reference-Guided Steering}
\label{app:response-numerics}
This appendix provides the implementation details of the Support Gate, Failure Gate, and Steering Coefficient introduced in Section~\ref{sec:method}.
All three use the temporal descriptor defined in Eq.~\eqref{eq:ordinal} and the reference banks retained after feature selection in Eq.~\eqref{eq:selected-feature}.
The banks contain descriptors of complete reference rollouts and remain fixed during deployment.
Feature selection requires at least two reference rollouts per outcome because its within-bank distances exclude self-pairs.

\subsection{Details of the Support Gate}
\label{app:support-details}
The Support Gate checks whether the current history provides sufficient evidence for comparison with the reference banks.
This subsection expands the Support Gate in Section~\ref{sec:method}.
We explain how the Dirichlet model generates the $M$ plausible temporal descriptors used to compute the support score $m_t$.
We specify how $\epsilon_{\mathrm{sup}}$ is determined offline to decide which reference descriptors count as nearby.

\textbf{Generating the descriptors.}
We use the triples $(y_i,y_{i+1},y_{i+2})$ and $(y_i,y_{i+2},y_{i+4})$ underlying the temporal descriptor in Eq.~\eqref{eq:ordinal}.
For each spacing, we count the occurrences of the six ordinal patterns within $y_{1:t}$ and add a pseudo-count of $1/2$ to each count to parameterize a Dirichlet distribution.
To generate each of the $M$ plausible descriptors, we draw one probability vector for each spacing.
Each vector contains six nonnegative probabilities that sum to one.
We compute its normalized entropy, $H(\mathbf p)=-\sum_{i=1}^{6}p_i\log p_i/\log 6$, and combine the six probabilities and entropy with the observed tie fraction and availability indicator.
Concatenating the two nine-dimensional summaries gives an 18-dimensional descriptor.
The tie fractions and availability indicators remain fixed across all $M$ descriptors.
Adjacent triples share activation values.
Since the Dirichlet model does not account for this dependence, its uncertainty estimates are approximate.

\textbf{Computing the support score.}
A sampled descriptor is supported if at least $k_{\mathrm{sup}}$ references in $\mathcal B_S\cup\mathcal B_F$ lie within raw Euclidean distance $\epsilon_{\mathrm{sup}}$.
The support score $m_t$ is the fraction of the $M$ sampled descriptors that are supported.
The support check passes when $m_t\geq\tau_m$ and both spacings meet the minimum triple-count requirement.
The Support Gate starts closed.
It opens after a fixed number of consecutive passing checks and closes after a fixed number of consecutive failing checks; otherwise, its state is retained.

\textbf{Setting the comparison radius.}
We determine $\epsilon_{\mathrm{sup}}$ from the reference banks before evaluation.
For each reference, we exclude the reference itself and measure its raw Euclidean distance to its $k_{\mathrm{sup}}$-th nearest neighbor in the pooled bank.
We set the radius to three times a prespecified quantile of these distances.
The quantile level lies strictly between zero and one.
The positive neighbor count $k_{\mathrm{sup}}$ cannot exceed the number of other references in the pooled bank.
The quantile level, neighbor count, and multiplier are fixed settings; the resulting radius depends on the reference data.
It is computed once and is not updated using evaluation rollouts.

\subsection{Details of the Failure Gate}
\label{app:distance-details}

The Failure Gate compares the observed descriptor $\mathbf v_t$ with the success and failure banks through $D_S(\mathbf v_t)$ and $D_F(\mathbf v_t)$.
The same success-bank distance also scores candidate corrections in Eq.~\eqref{eq:alpha-selection}.
We explain its calculation here so that both uses follow exactly the same comparison.

\textbf{Scaling descriptor coordinates.}
Unlike the Support Gate, the Failure Gate and coefficient selection in Eq.~\eqref{eq:alpha-selection} use scaled Euclidean distances.
We compute a scale for each of the 18 coordinates of the temporal descriptor in Eq.~\eqref{eq:ordinal}.
Each reference rollout contributes one descriptor to its outcome bank.
For coordinate $q$, we compute the mean $\mu_{S,q}$ and population variance $\sigma_{S,q}^{2}$ of its values across descriptors in the success bank.
We obtain $\mu_{F,q}$ and $\sigma_{F,q}^{2}$ in the same way from the failure bank.
These statistics therefore describe variation across reference rollouts, not across timesteps within a rollout.
Population variances use the corresponding bank size as the denominator.
To give success and failure equal weight when computing the scale, regardless of bank size, we set
\begin{equation}
s_q =
\max\!\left\{
10^{-8},
\sqrt{
\frac{\sigma_{S,q}^{2}+\sigma_{F,q}^{2}}{2}
+
\frac{(\mu_{S,q}-\mu_{F,q})^{2}}{4}
}
\right\}.
\label{eq:app-coordinate-scale}
\end{equation}
This scale accounts for both within-bank variation and differences between the bank means.
The floor prevents division by zero for constant coordinates.
All scales are computed offline and remain fixed during deployment.
For a query descriptor $\mathbf v$ and a reference descriptor $\mathbf u$, let $v_q$ and $u_q$ denote their values at coordinate $q$.
Their scaled Euclidean distance is
\begin{equation}
\operatorname{dist}(\mathbf v,\mathbf u)
=
\left[
\sum_{q=1}^{18}
\left(\frac{v_q-u_q}{s_q}\right)^2
\right]^{1/2}.
\label{eq:app-scaled-distance}
\end{equation}
The query may describe the observed execution prefix or a history modified by a candidate correction; both use the same fixed scales.

\textbf{Computing $D_S$ and $D_F$.}
Let $k$ be the requested number of neighbors for outcome comparison.
This setting is separate from $k_{\mathrm{sup}}$, which counts nearby references for the Support Gate.
We use $k_{\mathrm{eff}} =\min\{k,|\mathcal B_S|,|\mathcal B_F|\}$ to ensure that both outcome comparisons average over the same number of references.
To compute $D_S(\mathbf v)$, we find the $k_{\mathrm{eff}}$ nearest success references under Eq.~\eqref{eq:app-scaled-distance} and average their
distances to $\mathbf v$.
We compute $D_F(\mathbf v)$ identically using the failure bank.
The Failure Gate's instantaneous check passes when $D_F(\mathbf v_t)<D_S(\mathbf v_t)$; equal distances do not pass.

\textbf{Using the same distance for steering.}
In Eq.~\eqref{eq:alpha-selection}, each candidate coefficient produces a modified temporal descriptor.
We compute its distance to the success bank using the same scales and neighbor count as above.
Nearest neighbors are recomputed for every candidate; the search does not hold a single success reference fixed.
Only the Support Gate uses the descriptors sampled from the Dirichlet distribution.
The Failure Gate uses the observed descriptor, and coefficient selection uses the candidate-edited descriptors.
Support assessment and offline feature selection continue to use raw Euclidean distances; the coordinate scaling described here applies only to outcome comparison and candidate scoring.

\subsection{Details of the Steering Coefficient}
\label{app:local-response-details}

Equation~\eqref{eq:alpha-selection} compares candidate coefficients by estimating how each hidden-state edit changes the current feature intensity.
Here, we explain the approximation used in that equation and the numerical checks applied before using it.

\textbf{Predicting Edited Feature Intensity.}
For the selected $f^*$, we define the positive-activation fraction
\begin{equation}
\rho_t
=
\frac{1}{N_t}
\sum_{j=1}^{N_t}
\mathbf 1\!\left\{
\left[E(\mathbf h_{t,j})\right]_{f^*}>0
\right\},
\label{eq:positive-activation-fraction}
\end{equation}
where $\mathbf 1\{\cdot\}$ is the indicator function.
Thus, $\rho_t$ is the fraction of measured tokens whose activation contributes positively to the rectified intensity in Eq.~\eqref{eq:feature-intensity}.
Output of the frozen encoder $E$ in Eq.~\eqref{eq:sae-representation} corresponding to feature $f^*$ is denoted by
$\left[E(\mathbf h)\right]_{f^*}$, which can be written as
\begin{equation}
\left[E(\mathbf h)\right]_{f^*}
=
\mathbf w_{f^*}^{\top}\mathbf h+b_{f^*},
\end{equation}
where $b_{f^*}$ is the corresponding scalar bias.
For a candidate coefficient $\alpha$, as illustrated in Eq.~\eqref{eq:intervention}, the corresponding edit of each output is
\begin{equation}
\left[E\!\left(\mathbf h'_{t,j}(\alpha)\right)\right]_{f^*}
\approx
\left[E(\mathbf h_{t,j})\right]_{f^*}
+
\alpha\mathbf w_{f^*}^{\top}\mathbf d.
\label{eq:local-feature-response}
\end{equation}

The feature intensity in Eq.~\eqref{eq:feature-intensity} averages
the rectified activations across the encoder's outputs. Therefore, the predicted
current intensity for candidate coefficient $\alpha$ is
\begin{equation}
\widehat y_t(\alpha)
\approx
y_t+
\alpha\rho_t\mathbf w_{f^*}^{\top}\mathbf d.
\label{eq:unclipped-candidate-intensity}
\end{equation}

\textbf{Estimating the change in feature intensity.}
Under an affine encoder, moving a hidden-state vector by $\alpha\mathbf d$ changes the selected feature's pre-activation by
$\alpha\mathbf w_{f^*}^{\top}\mathbf d$.
If the SAE active set remains unchanged locally, tokens on which the feature is positive contribute this response, while inactive tokens contribute zero.
The intensity in Eq.~\eqref{eq:feature-intensity}
averages rectified activations across measured tokens.
When the measured and edited token sets are identical, averaging therefore gives the estimated intensity change
$\alpha\rho_t\mathbf w_{f^*}^{\top}\mathbf d$, where $\rho_t$ is the fraction of measured tokens on which the feature is positive.
This explains the predicted current intensity $\max\{0,y_t+\alpha\rho_t \mathbf w_{f^*}^{\top}\mathbf d\}$ inside Eq.~\eqref{eq:alpha-selection}.
The feature-dependent response is fixed offline; under the assumptions above, it is $\mathbf w_{f^*}^{\top}\mathbf d$.
Only the active-token fraction $\rho_t$ changes with the current hidden activations.
The dot-product expression is not an identity for arbitrary SAE encoders or edits that change which features are active.

\textbf{Evaluating candidate corrections.}
Each candidate replaces only the current intensity in the history used by $\phi$; earlier entries remain unchanged during candidate evaluation.
The resulting descriptor is compared with the success bank using $D_S$, as defined above.
This calculation evaluates a local change in feature statistics rather than predicting future actions or task success.
We abstain if the estimated intensity change per unit coefficient is nonfinite or its absolute value is below the configured minimum.
Otherwise, coefficient selection follows Eq.~\eqref{eq:alpha-selection}, and the chosen coefficient modifies the targeted hidden states through Eq.~\eqref{eq:intervention}.
As specified in the main text, steering is disabled whenever either gate is closed.

\textbf{Preprocessing feature intensities.}
The implementation retains raw intensities and applies $\widetilde y_t=\log(1+y_t/s_f)$ before constructing the temporal descriptor, where $s_f>0$ is a feature-specific scale fixed offline.
In exact arithmetic, this strictly increasing transformation preserves ordering and ties.
It therefore leaves the ordinal descriptor in Eq.~\eqref{eq:ordinal} unchanged.

The predicted intensity change is computed in the raw intensity space, before this transformation.
This keeps the response approximation consistent with the intensity measured in Eq.~\eqref{eq:feature-intensity}.

\section{Experimental Protocol and Hyperparameters}
\label{app:experimental-config}
\textbf{Simulated Fine-tuning and Evaluations.} Both $\pi_{0.5}$ and Fast-WAM
are fine-tuned on 500 successful demonstrations per task collected
by scripted policies.
All compared methods use the same fine-tuned checkpoint for a given
backbone and evaluation setting.
The action-model parameters remain frozen during steering and evaluation.

\textbf{Real-Robot Fine-tuning and Evaluations.} The real-robot $\pi_{0.5}$ policy is fine-tuned using 30 demonstrations
per task, and each method is evaluated over 30 trials per task.
These counts are separate from the simulation budgets above.

\textbf{Online reference and evaluation rollouts.}
Each task uses 24 outcome-labeled online
reference rollouts, with at least two successful and two failed episodes.
These policy rollouts are distinct from the scripted demonstrations
used for fine-tuning.
Evaluation uses 200 held-out online rollouts per task and method,
separate from the reference set.
SAE training uses hidden activations from the online rollout data
collected at a budget of 24 episodes per task.

\textbf{Shared SAE training.}
We train TopK sparse autoencoders on hidden activations pooled
across multiple tasks, using 24 online rollouts per task.
Each SAE uses an expansion factor of 8 and retains the 100 largest
encoder activations per token.
Writing $d$ for the hidden-state dimension and $K$ for the dictionary
size as in Section~\ref{sec:method}, these settings correspond to
$K=8d$ and $k_{\mathrm{SAE}}=100$.
The SAE is shared across tasks and remains frozen.
The TopK sparsity setting specifies the number of retained code
entries; task adaptation selects a decoder direction for steering.

\textbf{Shared Control Settings.}
Table~\ref{tab:steeract-hyperparameters} lists the control settings.
The same values are used across tasks.

\input{tables/hypermeters}

\section{Representation and Intervention-Site Ablations}
\label{app:ablation-variants}

We describe the variants in Table~\ref{tab:ablations}(a).
The alternative descriptors summarize different properties of
feature-intensity histories, including their distributions,
increments, event structure, and temporal dependence.
These are standard families of time-series statistics
\citep{fulcher2014highly}, but the specific combinations below
are constructed for our ablations rather than reproductions
of a single published method.

\textbf{Base.}
The original frozen action model executes without activation
steering. This reference measures whether each intervention
variant improves upon the unmodified policy.

\textbf{Marginals.}
This variant primarily summarizes the distribution of activation
intensities. Given the transformed intensity history
$x_i=\log(1+y_i/s_f)$, it concatenates histogram-bin frequencies,
quantiles, the mean, standard deviation, and median absolute
deviation. It also includes the fractions of measurements below
a low-intensity threshold and above a high-intensity threshold,
together with the fraction of near-zero one-step increments.
The representation therefore emphasizes activation levels and
dispersion, while retaining a coarse measure of local stability.
It does not encode the distribution of ordinal patterns.

\textbf{Increments.}
This variant represents changes in activation intensity at
multiple fixed lags. For each lag $\ell$, it forms
$\delta_i^{(\ell)}=x_{i+\ell}-x_i$ and summarizes these
increments using histogram-bin frequencies, quantiles,
the fractions of positive, negative, and near-zero changes,
and the mean absolute change.
A validity indicator records whether the prefix contains
enough measurements for that lag.
The lag-specific summaries are concatenated.
Unlike ordinal patterns, this representation explicitly
retains the magnitudes of activation changes.

\textbf{Events.}
This variant summarizes threshold-defined activation events.
Using fixed low and high intensity thresholds, it measures
the fraction of time above each threshold, activation-onset
frequency, and the mean and maximum lengths of consecutive
above-threshold runs.
Run lengths and event counts are normalized by prefix length.
Additional statistics describe near-zero increments, local
peak frequency and amplitudes, and the maximum and mean
intensity.
These events are defined from the activation signal itself,
not from manually annotated behavioral phases.
The representation emphasizes activation bursts, persistence,
and transitions rather than local ordinal-pattern frequencies.

\textbf{Autocorrelation.}
The variant labeled \emph{Autocorr.} combines autocorrelation
and spectral statistics.
It computes autocorrelations of the mean-centered history
at fixed lags, with indicators for valid estimates.
It also applies a Hann window and summarizes the resulting
Fourier power spectrum through normalized spectral entropy
and relative power in fixed frequency bands.
Additional statistics describe increment magnitude and
variability, turning-point frequency, and the mean and
standard deviation of intensity.
Thus, this variant captures repeated fluctuations and
longer-lag dependence; it is not an autocorrelation-only
representation.

\textbf{Raw NN.}
This control performs nearest-reference matching directly
on feature-intensity histories instead of comparing
fixed-dimensional statistical descriptors.
It tests whether sequence-level matching provides a
sufficient reference signal without summarizing the history
through ordinal patterns or other handcrafted statistics.

\textbf{Raw DTW.}
This control matches feature-intensity histories using
dynamic time warping (DTW)
\citep{sakoe1978dynamic}.
DTW computes a sequence-matching cost while allowing
monotonic, nonlinear temporal alignment between histories.
This comparison tests whether explicitly accommodating
differences in execution timing can replace matching
temporal-pattern summaries.

\textbf{Enc and Dec.}
These variants use the ordinal-pattern representation
underlying \method{}, based on local orderings and
permutation entropy \citep{bandt2002permutation}.
Enc intervenes in the action model's context branch,
whereas Dec intervenes in its action expert.
Both use the reference-guided steering formulation;
the comparison evaluates the effect of the intervention
site. Enc is the default single-feature configuration
reported in the main experiments.

\section{Qualitative Paired Simulation Rollouts}

Figures~\ref{fig:paired-sim} and ~\ref{fig:paired-sim-1} jointly
present a single collection of paired qualitative comparisons on
RoboTwin~2.0 and DexJoCo.
For each task, the upper row executes the frozen backbone without
intervention (\emph{No steering}), while the lower row applies
\method{} from a matched initial scene. Frames progress from left
to right and summarize the resulting closed-loop executions. These
pairs are selected failure-to-success examples intended to expose
behavioral differences.

\section{Limitations.} 
Our evidence that intervention timing matters
comes from gate and fixed-delay ablations, not from a controlled
experiment that isolates behavioral phase from instantaneous state.
Gains do not close backbone-dependent gaps (e.g., Hanoi Bi remains at
0\%), and the method requires hidden-state access and a pretrained SAE
whose quality bounds the available steering directions.

\input{figures/figure_sim_examples}
\input{figures/figure_sim_examples_1}

%% file: tables/tab-7.tex
\begin{table}[tbp]
\centering
\caption{Ablations on RoboTwin~2.0 with $\pi_{0.5}$:
(a) temporal descriptors, matching rules, feature counts,
steering strengths, and intervention sites;
(b) total source rollouts; (c) successful source rollouts. (d) feature count and coefficient adaptation.
Entries are success rates (\%, $\uparrow$); Mean averages over the nine displayed tasks.
Matching controls use Enc and retain the ordinal-based support gate.
Gray highlights the full method.
\textbf{Bold}/\underline{underline} mark the best/second-best distinct results per column within each group, including ties.}
\label{tab:ablations}
\footnotesize
\setlength{\tabcolsep}{3pt}
\renewcommand{\arraystretch}{1.12}
\resizebox{\linewidth}{!}{%
\begin{tabular}{@{}l*{10}{c}@{}}
\toprule
Method / Setting
& \shortstack{Put Bottles\\in Dustbin}
& \shortstack{Stack Three\\Blocks}
& \shortstack{Rank Blocks\\by Size}
& \shortstack{Place Dual\\Shoes}
& \shortstack{Dump Bin\\into Big Bin}
& \shortstack{Open\\Microwave}
& \shortstack{Turn\\Switch}
& \shortstack{Place\\Mouse Pad}
& \shortstack{Pick Diverse\\Bottles}
& Mean \\
\midrule

\multicolumn{11}{@{}l}{
\textit{(a) Representation and steering ablations}} \\
\midrule
Base
& 74 & 72 & 43 & 73 & 89 & 84 & 47 & 62 & 77
& 69.0 \\

Marginals
& 76 & 76 & 48 & 77 & 88 & 77 & 49 & \underline{67} & 82
& 71.1 \\

Increments
& 78 & 76 & 39 & 78 & 89 & 83 & 43 & 61 & \underline{83}
& 70.0 \\

Events
& 76 & 78 & 45 & 77 & 89 & 82 & \underline{50} & 54 & 80
& 70.1 \\

Autocorrelation
& \underline{82} & \underline{83}
& 49 & \textbf{83} & \underline{93} & 86 & 49 & 56 & \textbf{84}
& \underline{73.9} \\

Raw NN
& 75 & 81 & \underline{52} & 76 & 87 & 89 & \textbf{52} & 66 & 72
& 72.2 \\

Raw DTW
& 78 & \textbf{84} & 49 & 62 & 92 & \textbf{96}
& \underline{50} & 66 & 68
& 71.7 \\

\method{} (Dec)
& 79 & \underline{83} & 39 & 81 & 92 & 87 & 44 & \underline{67} & 66
& 70.9 \\

\rowcolor[gray]{0.93}
\method{} (Enc)
& \textbf{88} & \underline{83}
& \textbf{58} & \underline{82} & \textbf{94} & \underline{93}
& \textbf{52} & \textbf{69} & 80
& \textbf{77.7} \\

\midrule
\multicolumn{11}{@{}l}{
\textit{(b) Total source rollouts --- \method{} (Enc)}} \\
\midrule
12 rollouts
& \underline{77} & 79 & 53 & \underline{79}
& \underline{93} & \underline{95} & 44 & 64 & 78
& \underline{73.6} \\

48 rollouts
& \textbf{79} & \underline{84}
& \textbf{63} & \textbf{84} & \textbf{94} & \textbf{96}
& \underline{47} & \underline{70} & \underline{79}
& \textbf{77.3} \\

96 rollouts
& 74 & \textbf{87}
& \underline{57} & \underline{79} & \textbf{94} & 94
& \textbf{51} & \textbf{74} & \textbf{86}
& \textbf{77.3} \\

\midrule
\multicolumn{11}{@{}l}{
\textit{(c) Successful source rollouts --- \method{} (Enc)}} \\
\midrule
8 successful rollouts
& \underline{77} & 79
& \underline{60} & 75 & \underline{90} & 88
& \textbf{54} & 66 & 81
& 74.4 \\

24 successful rollouts
& \textbf{79} & \underline{86}
& \textbf{64} & \underline{77} & \textbf{94} & \underline{94}
& 49 & \textbf{71} & \underline{84}
& \underline{77.6} \\

48 successful rollouts
& 76 & \textbf{87} & 59 & \textbf{80} & \textbf{94} & \textbf{96}
& \underline{52} & \underline{70} & \textbf{87}
& \textbf{77.9} \\

\midrule
\multicolumn{11}{@{}l}{
\textit{(d) Feature count and coefficient adaptation --- \method{} (Enc)}} \\
\midrule
Top-$2$ features
& \textbf{90} & 87 & 59 & \textbf{84} & \underline{94}
& \underline{96} & \underline{51} & \textbf{73} & \underline{82}
& \underline{79.6} \\

Top-$4$ features
& \textbf{90} & \textbf{89} & 62 & \underline{81} & \textbf{95}
& \underline{96} & \textbf{54} & \underline{71} & \underline{82}
& \textbf{80.0} \\

Fixed $\alpha=1$
& \underline{86} & 81 & \textbf{66} & 77 & 90
& 93 & 44 & 66 & 79
& 75.8 \\

Fixed $\alpha=2$
& \underline{86} & \underline{88} & \underline{63} & 68 & 86
& \textbf{97} & 49 & 59 & \textbf{85}
& 75.7 \\

\bottomrule
\end{tabular}%
}
\end{table}

%% file: tables/tab-8.tex
\begin{table}[tbp]
\centering
\caption{Timing and gating comparisons on VLABench.
Success rates (\%, $\uparrow$) use 200 evaluation episodes
per task and reported setting; Mean is the task macro-average.
Gray highlights maturity-based steering.
\textbf{Bold}/\underline{underline} indicate the best/second-best
distinct reported results per column, including ties.}
\label{tab:pilot-timing}
\footnotesize
\setlength{\tabcolsep}{3pt}
\renewcommand{\arraystretch}{1.12}
\resizebox{0.75\linewidth}{!}{%
\begin{tabular}{@{}l*{9}{c}@{}}
\toprule
Method
& Painting
& Book
& Drinking
& \shortstack{Chemistry\\Tube}
& Poker
& Mahjong
& Toy
& Fruit
& Mean \\
\midrule
Frozen
& \underline{17} & 46 & 15 & 47
& \underline{18} & 6 & 7 & \underline{33}
& 23.63 \\


Delay 10
& \underline{17} & \textbf{51} & 22 & 47
& 13 & \underline{13} & 5 & 29
& 24.63 \\

Delay 20
& \underline{17} & 44 & 19 & \underline{52}
& 16 & 11 & \underline{12} & 28
& \underline{24.88} \\

Without support gate
& 15 & 49 & \underline{23} & 49
& 12 & 8 & 5 & 24
& 23.13 \\

Without failure gate
& 13 & 48 & 27 & 53
& 22 & 18 & 25 & 33
& 29.88 \\

Without gates
& 17 & 43 & 24 & 42
& 22 & 9 & 3 & 19
& 22.38 \\

\rowcolor[gray]{0.93}
\method{}
& \textbf{34} & \underline{50} & \textbf{39} & \textbf{60}
& \textbf{22} & \textbf{14} & \textbf{26} & \textbf{44}
& \textbf{36.13} \\
\bottomrule
\end{tabular}%
}
\end{table}

%% file: figures/algorithm.tex
\begin{algorithm}[t]
\caption{\method{}: Offline Selection and Online Steering}
\label{alg:steeract}
\small
\begin{algorithmic}[1]
\Require Frozen action model, pretrained SAE,
outcome-labeled rollouts, candidate features $\mathcal C$,
and fixed control settings.

\Statex \textbf{Offline: select a feature and its references}
\State For each $f\in\mathcal C$, construct banks
$\mathcal B_f^S,\mathcal B_f^F$ from descriptors 
in Eq.~\eqref{eq:feature-intensity} and Eq.~\eqref{eq:ordinal}.
\State Select $f^*$ and its decoder direction $\mathbf d$
using Eq.~\eqref{eq:selected-feature}.
\State Retain $\mathcal B_S,\mathcal B_F$ for $f^*$;
calibrate the support radius and coordinate scales
(Appendix~\ref{app:response-numerics}).

\Statex \textbf{Online: steer a new rollout}
\State Initialize empty history, $\alpha_0\gets0$,
and closed gates with zeroed counters.
\For{each robot decision $t$ until termination}
    \State Run the model to the intervention site;
    obtain $\mathbf Z_t$ and the unedited intensity $y_t$
    using Eqs.~\eqref{eq:sae-representation}
    and~\eqref{eq:feature-intensity}.
    \State Append $y_t$; compute
    $\mathbf v_t=\phi(y_{1:t})$
    using Eq.~\eqref{eq:ordinal},
    retaining the pattern counts.
    \State Compute $m_t$ using Sec.~\ref{sec:online-steering};
    update the temporal support gate using
    $m_t\geq\tau_m$.
    \State If support is open, update the failure gate
    using Sec.~\ref{sec:online-steering};
    otherwise reset it and its counters.

    \State $\alpha_t\gets0$.
    \If{both gates are open}
        \State $\alpha_t\gets
        \Call{BoundedSearch}
        {y_{1:t},\mathbf Z_t,\alpha_{t-1}}$,
        evaluating candidates with
        Eq.~\eqref{eq:alpha-selection}.
    \EndIf

    \If{$\alpha_t\neq0$}
        \State Edit each targeted token
        using Eq.~\eqref{eq:intervention}.
        \State Re-encode the edited states and replace $y_t$
        with the measured intensity
        from Eq.~\eqref{eq:unclipped-candidate-intensity}.
    \EndIf
   \State  Retain the applied $\alpha_t$ for the next decision.
    \State Resume the model from the resulting hidden states
    and execute its action output.
\EndFor
\end{algorithmic}
\end{algorithm}

%% file: tables/hypermeters.tex
\begin{table}[htbp]
\centering
\caption{Shared \method{} control hyperparameters.}
\label{tab:steeract-hyperparameters}
\small
\begin{tabular}{@{}llr@{}}
\toprule
Parameter & Symbol & Value \\
\midrule
Support samples & $M$ & 500 \\
Support-gate threshold & $\tau_m$ & 0.65 \\
Maximum coefficient magnitude & $A_{\max}$ & 10 \\
Coefficient grid spacing & $\delta_\alpha$ & 0.5 \\
\bottomrule
\end{tabular}
\end{table}

%% file: figures/figure_sim_examples.tex
\begin{figure}[H]
\centering
\includegraphics[width=\linewidth]{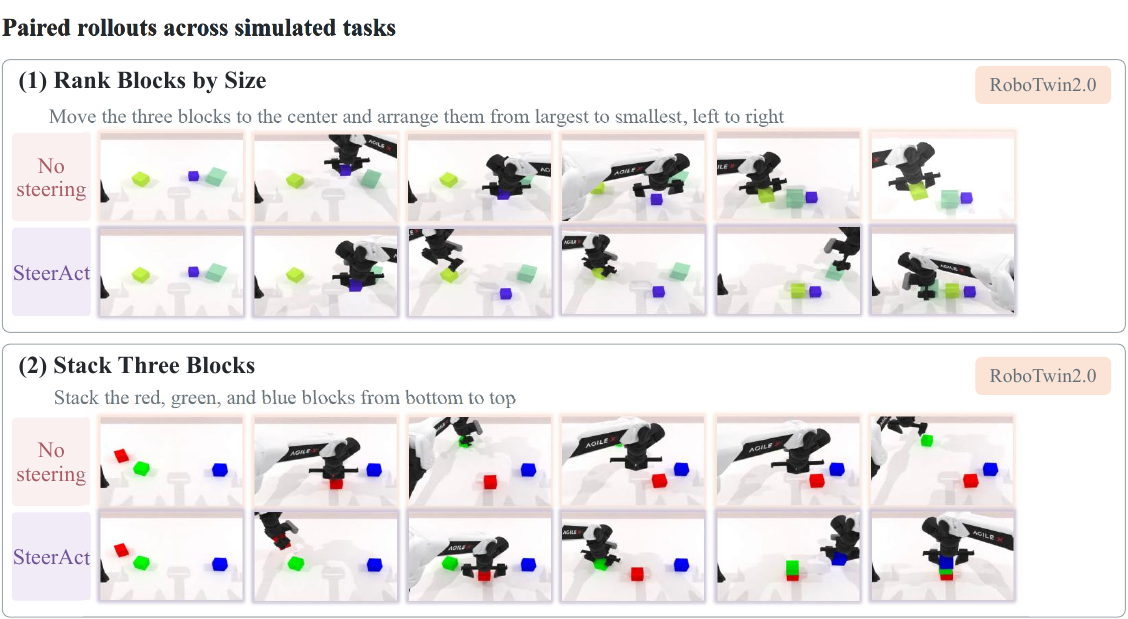}
\includegraphics[width=\linewidth]{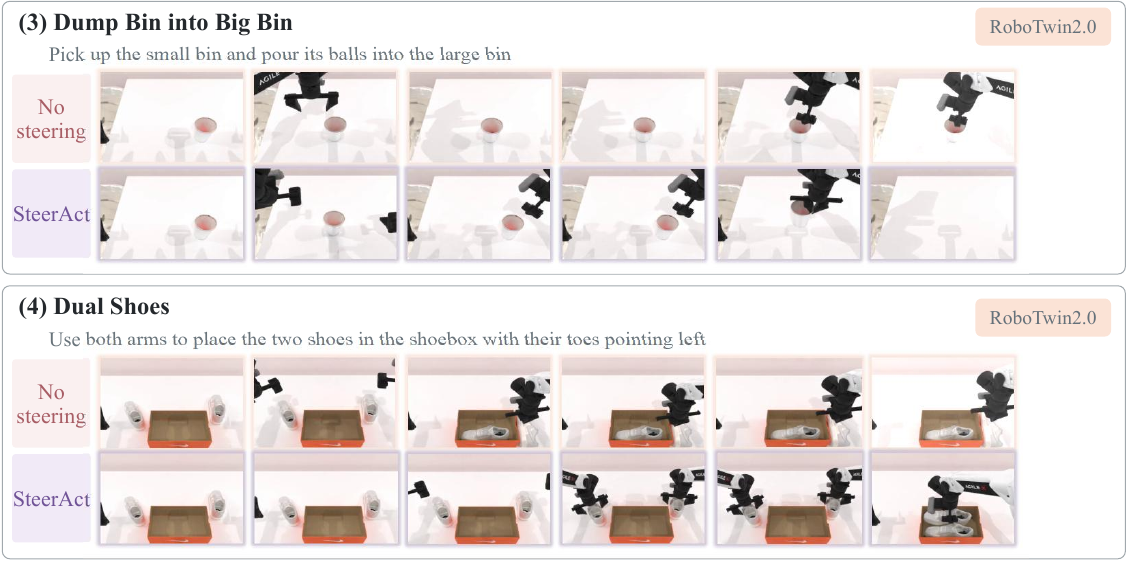}
\caption{Simulated rollouts on RoboTwin~2.0.
Each panel compares the unsteered frozen backbone (top) with
\method{} (bottom) from a matched initial scene; time advances
from left to right. }
\vspace{-13pt}
\label{fig:paired-sim}
\vspace{-13pt}
\end{figure}

%% file: figures/figure_sim_examples_1.tex
\begin{figure}[H]
\centering
\includegraphics[width=\linewidth]{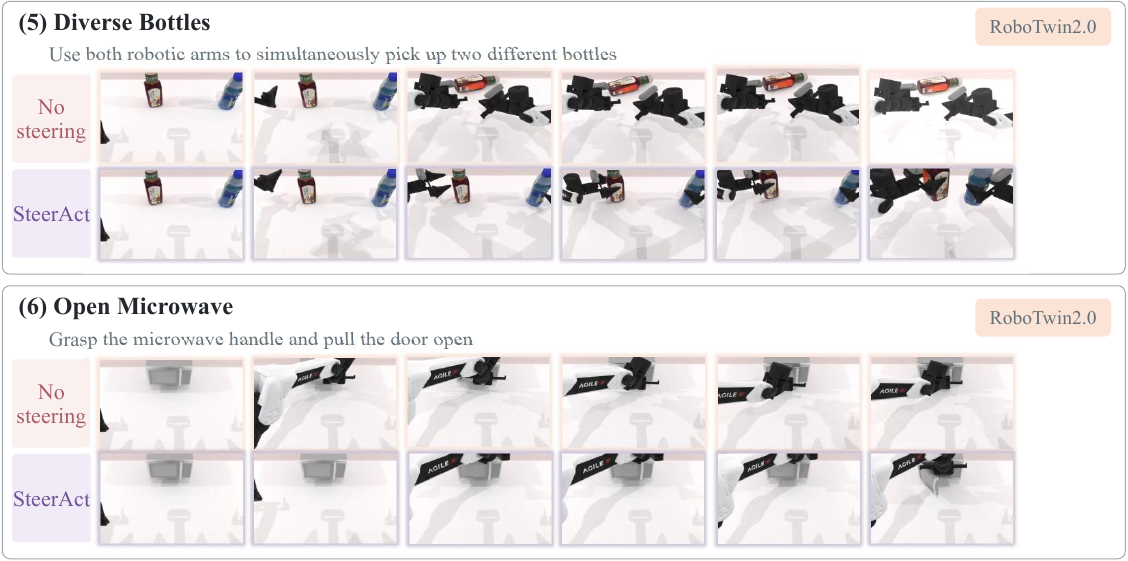}
\includegraphics[width=\linewidth]{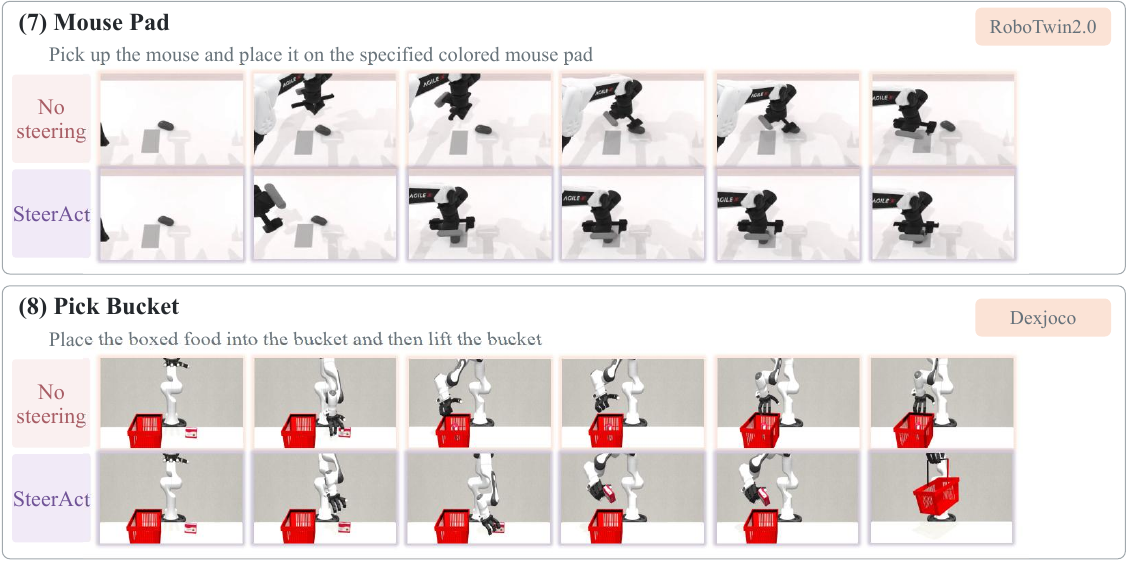}
\includegraphics[width=\linewidth]{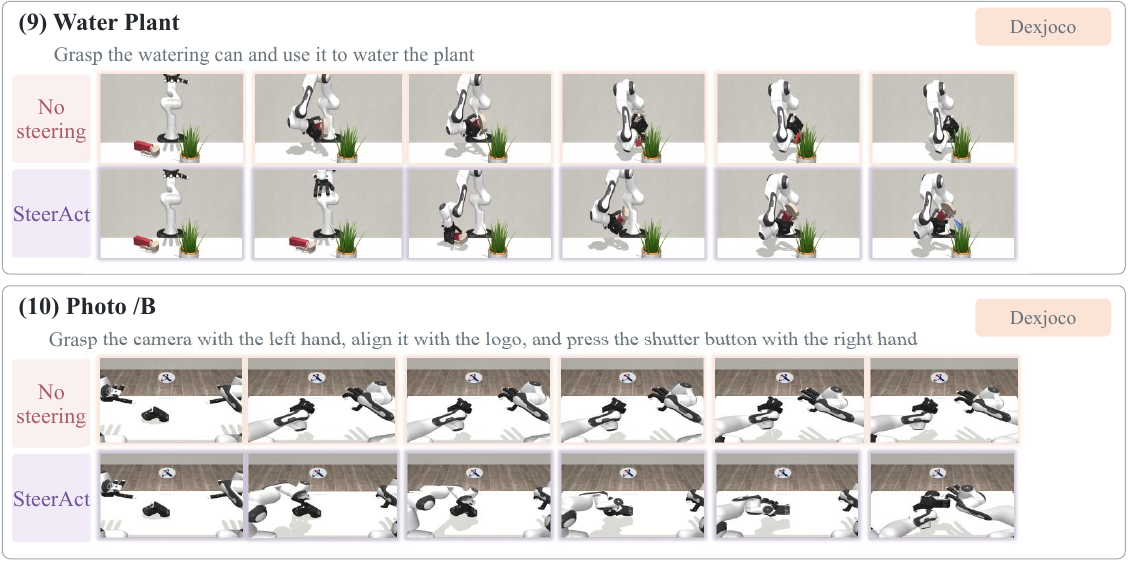}
\caption{Paired simulated rollouts on RoboTwin~2.0 and DexJoCo (continued). The unsteered and steered rows begin from matched initial scenes, and frames advance from left to right.}
\vspace{-13pt}
\label{fig:paired-sim-1}
\vspace{-13pt}
\end{figure}